\documentclass[11pt]{article}
\usepackage{fullpage}

\usepackage{graphicx}  
\usepackage{amssymb}
\usepackage{amsmath}  
\usepackage{amstext}  
\usepackage{amsthm}   
\usepackage{natbib}
\usepackage[flushleft]{threeparttable}  
\usepackage{float}  
\usepackage{hyperref}  
\newtheorem{definition}{Definition}
\usepackage{algorithm,algorithmic}
\newtheorem{thm}{Theorem}

\newtheorem{ass}{Assumption}

\hypersetup{colorlinks = true, citecolor = blue}  
\usepackage{subfigure}

\def \E {\mathrm{E}}
\def \g {\mathbf{g}}

\def \R {\mathbb{R}}

\def \N {\mathcal{N}}

\def \P {\mathcal{P}}

\def \E {\mathrm{E}}

\def \g {\mathbf{g}}
\def \R {\mathbb{R}}

\def \Lh {\widehat{\mathcal L}}
\def \Ls {\mathcal L}

\def \rh {\widehat{r}}
\def \G {\mathcal{G}}

\begin{document}
\title{\bf A Theoretical Analysis of Learning with Noisily Labeled Data}
\author{
Yi Xu, Qi Qian, Hao Li, Rong Jin\\
Machine Intelligence Technology, Alibaba Group\\
\{yixu, qi.qian, lihao.lh, jinrong.jr\}@alibaba-inc.com
}
\date{First Version: \today 
}
\maketitle

\begin{abstract}
Noisy labels are very common in deep supervised learning. Although many studies tend to improve the robustness of deep training for noisy labels, rare works focus on theoretically explaining the training behaviors of learning with noisily labeled data, which is a fundamental principle in understanding its generalization. In this draft, we study its two phenomena, clean data first and phase transition, by explaining them from a theoretical viewpoint. Specifically, we first show that in the first epoch training, the examples with clean labels will be learned first. We then show that after the learning from clean data stage, continuously training model can achieve further improvement in testing error when the rate of corrupted class labels is smaller than a certain threshold; otherwise, extensively training could lead to an increasing testing error.
\end{abstract}

\section{Introduction}
Noisy labels are very common in the real world~\citep{xiao2015learning,li2017webvision,lee2018cleannet,song2019selfie} due to some reasons such as data labeling by human mistake and the complexity of label.
In many tasks, even if wrongly labeled examples could be recognized by a human expert, it is impossible to manually review all training examples~\citep{koh2017understanding}. Learning with noisily labeled data is probably inevitable and becomes a common challenge in supervised learning, especially for large-scale deep learning problems.

It has been shown that noisy labels would lead to performance degradation when training deep neural networks (DNNs) because of over-fitting to the noise~\citep{arpit2017closer}.
However, it is a fundamental principle for understanding the generalization of deep learning models~\citep{zhang2016understanding}. 
Empirically, learning with noisily labeled data has been examined by many recent studies such as~\citep{song2019selfie,lyu2019curriculum,wang2019symmetric,li2020dividemix,liu2020early,nguyen2020self}. Most of them are designed to improve learning performance by make a supervised learning process more robust to label noise~\citep{song2020learning,algan2021image}. Rare studies focus on explaining the training behavior of noisily labeled data from the view of theory. {Empirical results have shown that even with a significant portion of the training data assigned to random labels, we are still able to learn an appropriate prediction model that can make accurate prediction~\citep{li2020dividemix,liu2020early}.} Two interesting phenomena have been observed from the task of learning from noisily labeled data. The first observation is {\it\bf clean data first}, i.e. data points with clean labels will be learned first. Although it was shown ``intuitively'' through a linear binary classification case~\citep{liu2020early}, the full story behind this belief is unclear. In the first epoch, we can consider the standard optimization analysis due to the statistical independence property of current model and sampled example(s)\footnote{We could use the uniform sampling strategy without replacement when implementing stochastic gradient descent during the training process.}. The convergence results shows that stochastic gradient descent (SGD)  essentially converges to the objective function of examples with clean labels in the order of $\widetilde O(1/m)$ \footnote{$\widetilde O(\cdot)$ suppresses a logarithmic factor}, where $m$ is the number of training examples with clean labels. The second observation is {\it\bf phase transition}, i.e. when $\gamma$, the percentage of data with corrupted class labels, is smaller than a certain threshold, continuously training model can lead to a further reduction in testing error after the stage of learning from clean data. However, when $\gamma$ exceeds the certain threshold, extensively training could result in an increase in testing error. In this draft, we aim to explain both phenomenon from a theoretical viewpoint, which could be considered as a theoretical understanding of learning with noisy labels.

\section{Preliminaries and Notations}
We introduce preliminaries in this section. We first give some notations and the problem definition,then we present some assumptions that will be used in the convergence analysis.

Let $\nabla_\theta f(\theta)$ denote the gradient of a function $f(\theta)$. When the variable to be taken a gradient is obvious, we use $\nabla f(\theta)$ for simplicity. We use $\|\cdot\|$ to denote the Euclidean norm. Let $\langle \cdot,\cdot\rangle$ be the inner product.

The classification problem aims to seek a classifier to map an instance $x\in\R^d$ onto one of labels $y\in\{+1,-1\}$. Suppose the instances are draw from a distribution $\P$, i.e., $x \sim \P$, where $\P$ is usually unknown. For simplicity, we assume $y = y(x)$, i.e. the class label of instance $x$ is decided by a deterministic function $y(x)$. Our goal is to learn a prediction function $f(x; \theta): \R^d\times\R^d \to \R$ that is as close as possible to $y$, where $\theta\in\R^d$ is the model parameter. To achieve this goal, traditional method is to minimize the following expected loss $\min_{\theta\in\R^d} \E_{x\sim\P}[1/2\|f(\theta;x)-y(x)\|^2]$, which is also known as expected risk minimization. Since the distribution $\P$ is unknown, solving the risk minimization is difficulty. However, one can sample $n$ examples $\{(x_1,y_1), \dots, (x_n,y_n)\}$ from $\P$ independently, then we aim to minimize an empirical loss $\min_{\theta\in\R^d} \frac{1}{2n}\sum_{i=1}^{n}\|f(\theta;x_i)-y_i(x_i)\|^2$, which is also known as empirical risk minimization. In this study, the training data set contains clean labels and noisy labels, we denote by $\gamma$, the percentage of data points with noisy labels. We divide the training data into two parts: 
\begin{itemize}
\item {\it clean data}: $D_a = \{(x^a_i, y^a_i), i=1,\ldots, m\}$, where $x_i^a \in \R^d$ is sampled independently from a given distribution $\P$ and $y_i^a = y(x_i^a)$ is a clean class label decided by function $y:\R^d \mapsto \{-1, +1\}$, and 
\item {\it corrupted data}: $D_b = \{(x^b_i, y^b_i), i=1, \ldots, n\}$, where $x_i^b$ is sampled independently from $\P$ and $y_i^b$ is assigned randomly with equal chance to be either $-1$ or $+1$.
\end{itemize}
Please note that the labels $y_i^b$ from corrupted data is independent of instance $x_i^a$, whose expectation is $0$, i.e. $\E[y_i^b] = 0$. 
The percentage of noisily labeled data points is $\gamma = n/(m+n)$. We consider the following expected risk to be optimized
\begin{align}\label{opt:rm}
   \Ls(\theta) = (1-\gamma)\Ls_a(\theta) + \gamma \Ls_b(\theta),
\end{align}
where $\Ls_a(\theta)$ is the loss function for clean labeled data and $\Ls_b(\theta)$ is the loss function related to noisily labeled data, whose definitions are given by
\begin{align}\label{opt:rm:loss}
    \Ls_a(\theta) = \E_{x\sim\P}\left[\frac{1}{2}|f(x;\theta)-y(x)|^2\right], \quad \Ls_b(\theta) = \E_{x\sim\P}\left[\frac{1}{2}|f(x;\theta)|^2\right].
\end{align}
In practice, we want to solve the following empirical risk minimization problem
\begin{align}\label{opt:erm}
    \min_{\theta\in\R^d}\Lh(\theta) := (1-\gamma)\Lh_a(\theta) + \gamma \Lh_b(\theta),
\end{align}
where $\Lh_a(\theta)$ is the loss function over training data with clean labels and $\Lh_b(\theta)$ is the loss function over training data with noisy labels, whose definitions are given by
\begin{align}\label{opt:erm:loss}
    \Lh_a(\theta) = \frac{1}{2m}\sum_{i=1}^{m}|f(x_i^a;\theta)-y_i^a|^2, \quad \Lh_b(\theta) = \frac{1}{2n}\sum_{i=1}^{n}|f(x_i^b;\theta)-y_i^b|^2.
\end{align}
{Since $\E[\Lh_b(\theta)] = \Ls_b(\theta) + \frac{1}{2}$, we may consider $\Ls(\theta)$ as the final objective function.} As a widely used method, SGD can be employed to solve problem~(\ref{opt:erm}) and it updates solutions iteratively by
\begin{align}\label{update:sgd}
    \theta_{t+1} = \theta_t - \eta g_t
\end{align}
where $t$ is the iteration number, $\eta>0$ is a learning rate, and $g_t$ is the stochastic gradient given by
\begin{align}\label{stoc:grad}
   g_t =  (f(x_{i_t};\theta_t)-y_{i_t}) \nabla_\theta f(x_{i_t};\theta_t)
\end{align}
with $(x_{i_t}, y_{i_t})$ is uniformly sampled from $D_a\cup D_b$.

To establish the theoretical analysis, we introduce some commonly used assumptions in the literature of optimization~\citep{ghadimi2013stochastic,xu2019non}. 
\begin{ass}\label{ass:1}
Assume the following conditions hold: 
\begin{itemize}
\item[(i)]  There exists two constants $F>0$ and $G>0$, such that $$\|f(x;\theta)\| \leq F, \;\|f(x;\theta) - y(x)\| \leq F,$$ and
$$\|(f(x;\theta) - y(x))\nabla f(x;\theta)\| \leq G, \; \|f(x;\theta)\nabla f(x;\theta)\| + \|\nabla f(x;\theta)\| \leq G.$$
\item[(ii)] The loss function $\Ls$ is smooth with an $L$-Lipchitz continuous gradient, i.e., it is differentiable and there exists a constant $L>0$ such that $$\Ls(\theta')\le \Ls(\theta'')+ \langle \nabla \Ls(\theta''), \theta' - \theta'' \rangle + \frac{L}{2}\|\theta' - \theta''\|^2, \forall \theta', \theta'' \in \R^d.$$ 
\end{itemize}
\end{ass}
We now introduce an important functional property, which is called Polyak-\L ojasiewicz (PL) condition~\citep{polyak1963gradient}. PL condition has been observed in training deep neural networks~\citep{allen2019convergence, yuan2019stagewise}, and it is widely used to establish convergence in the literature of optimization~\citep{yuan2019stagewise, wang2019spiderboost, karimi2016linear, li2018simple, charles2018stability, li2020exponential}. 
\begin{definition}\label{def:PL} 
    A function $\Ls$ satisfies $\tau$-PL condition if there exists a constant $\tau>0$ such that $2\tau(\Ls(\theta)- \Ls^*) \le \|\nabla \Ls(\theta)\|^2, \forall \theta\in\R^d$, where $\Ls^* = \min_{\theta\in\R^d} \Ls(\theta)$ is the optimal value.
\end{definition}
For the simplicity, we set $\Ls_a^{min} := \min_{\theta\in\R^d} \Ls_a(\theta) = 0$ and $\Ls_b^{min} := \min_{\theta\in\R^d} \Ls_b(\theta) = 0$ in our proofs without loss of generality, which is a common property observed in training deep neural networks~\citep{zhang2016understanding,allen2019convergence,du2018gradient,du2019gradient,arora2019fine,chizat2019lazy, hastie2019surprises,yun2019small, zou2020gradient}.
\section{Main Results}
We first study the observation of learning from data with clean labels first, and then the observation of phase transition in terms of $\gamma$.
\subsection{Learning from data with clean labels first}
To this end, we will focus on the optimization of the first epoch. Note that in the first epoch, it satisfies the condition of statistical independence and therefore we can apply the standard analysis of stochastic gradient descent. 
\begin{thm}\label{thm:1:first:stage}
Under Assumption~\ref{ass:1}, suppose $\Ls_a$ and $\Ls_b$ satisfies $\mu_a$-PL condition and $\mu_b$-PL condition, respectively. 
Let $\theta_0$ be our initial solution, and $\theta_1$ be the solution after running through the first epoch of data, and $\tau := \min(2\mu_a, 2\mu_b)$. By selecting $\eta  \leq 1/L$ and $m \geq \frac{4G^2 L}{\tau^2\Ls(\theta_0)}$.
We have 
\[
\E\left[\Ls_a(\theta_1)\right] \leq O\left(\frac{4G^2\log m}{(1-\gamma)\tau^2m}\right).
\]
\end{thm}
{\bf Remark.} Through the first epoch of optimization, since we do not reuse any data point, either clean or corrupted one, SGD essentially optimizes the objective function $\Ls(\theta)$, instead of $\Ls_a(\theta)$. As indicated by the above analysis, despite of the presence of noisy labels, the resulting solution is still able to reach a low value of $\Ls_a(\theta)$, which is on the order of $O(\log(m)/m)$. Of course, the presence of noisy labels indeed hinders the progress of optimization, through parameter $\gamma$ (i.e. the percentage of noisily labeled data), $G^2$ and $\tau$. This analysis explains why for learning from noisily labeled data, it appears that the model could learn clean data first because for early epoch, noisily labeled data behave like regularizer to our solution.

\subsection{Phase Transition for $\gamma$}
Now, let’s examine the late stage of the optimization beyond the first epoch, in which we do not have the statistical independence between the model to be updated and the training data. As a result, our objective function ``becomes" $\Lh(\theta) = (1-\gamma)\Lh_a(\theta) + \gamma \Lh_b(\theta)$. Let $\theta_1$ be the solution obtained from the first epoch.
Define
\[
    \Omega(\rh,s) = \left\{\theta: \frac{1}{m}\sum_{i=1}^m\left\|f(x_i^a;\theta) - y(x_i^a)\right\|^2 \leq \rh^2, \frac{1}{m}\sum_{i=1}^m\left\|f(x_i^a;\theta) - y(x_i^a)\right\| \leq \sqrt{s} \rh\right\}
\]
where
\[
\rh^2 := \frac{1}{m}\sum_{i=1}^m\left\|f(x_i^a;\theta_1) - y(x_i^a)\right\|^2,\quad \sqrt{s} := \frac{1}{m\rh}\sum_{i=1}^m\left\|f(x_i^a;\theta_1) - y(x_i^a)\right\|
\]
We assume that $s \ll m$, implying that vector $(f(x_1;\theta), \ldots, f(x_m;\theta))$ is a sparse vector. We first give the following results of bounding the variances of the stochastic gradients, which will be used in our analysis of the main result.
\begin{thm}\label{thm:main:1:max:norm}
Under the assumption $m \geq 64F^2\log(2/\delta)$, we have, with a probability $1 - 3\delta$, 
\begin{align}\label{thm:main:1:max:norm:L:a}
&\max\limits_{\theta \in \Omega(\rh, s)} \| \nabla \Ls_a(\theta) - \nabla \Lh_a(\theta) \| 
\leq  C\left(\frac{GF\log(1/\delta)}{m} + G\rh\left[\sqrt{\frac{\log(1/\delta)}{m}} + \sqrt{\frac{s}{m}\log\frac{2m}{s}}\left(1 + \log m\right) \right] \right),\\
&\max\limits_{\theta \in \Omega(\rh, s)} \| \nabla \Ls_b(\theta) - \nabla \Lh_b(\theta) \| 
\leq  C\left(\frac{GF\log(1/\delta)}{n} + G F\left[\sqrt{\frac{\log(1/\delta)}{n}} + \sqrt{\frac{s}{n}\log\frac{2n}{s}}\left(1 + \log n\right) \right] \right),
\end{align}
where $C$ is a universal constant. 
\end{thm}
To understand the phenomenon of phase transition (i.e. the optimizer may continuously reduce the testing error when $\gamma$ is smaller than a threshold, and will lead to an increase in testing error after extensive training if $\gamma$ is above the threshold), we will examine the condition for all the solutions in the later iterations after the first epoch that will stay within $\Omega(\rh, s)$. If solution $\theta$ continuously stays in $\Omega(\rh, s)$, we will not see the degradation in performance. Otherwise, we may find a large error after the training of the first epoch.  To this end, we will assume that is the case, and see when the condition will be violated. {In this case, the stochastic gradient $g_t$ is not a unbiased estimator of $\nabla\Ls(\theta'_t)$, i.e. $\E[g_t]\neq \nabla\Ls(\theta'_t)$
}
\begin{thm}\label{thm:2:after:first:stage}
Under Assumption~\ref{ass:1}, suppose $\Ls_a$ and $\Ls_b$ satisfies $\mu_a$-PL condition and $\mu_b$-PL condition, respectively. 
Let $\theta_1$ be the solution after running through the first epoch of data, and $\tau := \min(2\mu_a, 2\mu_b)$. By selecting $\eta = \frac{2}{\tau m}\log\left( \frac{m\tau^2\Ls(\theta_0)}{8G^2 L}\right)$ and $m \geq \frac{2L}{\tau }\log\left( \frac{m\tau^2\Ls(\theta_0)}{8G^2 L}\right)$. If we have
\begin{align}\label{condition:phase:transition}
    1-\gamma \geq\max\left\{ \frac{C''G^2\Delta (\rh^2 + F^2)}{8\tau\rh^2(n+m)}, \frac{16G^2L\log\left(\frac{\Ls(\theta_1)}{\rh^2(1-\gamma)}\right)}{\tau^2 \rh^2 t}\right\}
\end{align}
where $C''$ is a universal constant, $\Delta := \left(\log\left(\frac{6}{\delta}\right)+s\log\left(\frac{2\max(m,n)}{s}\right)\log(\max(m,n))\right)$, then
\begin{align}\label{phase:transition}
\E\left[\Ls_a(\theta'_t)\right] \leq \frac{\rh^2}{2}.
\end{align}
\end{thm}
{\bf Remark.} The theorem shows that when (\ref{condition:phase:transition}) holds, the solution $\theta$ continuously stays in $\Omega(\rh, s)$ and we will not see the degradation in performance. The condition (\ref{condition:phase:transition}) essentially leads to the so called phase transition phenomenon, i.e. the performance over the testing data would degrade with more training epochs. In addition, if the condition (\ref{condition:phase:transition}) does not hold, then we may have a worse performance over the testing data after the first epoch.

\section{Proofs}
In this section, we provide the proofs of theorems in previous section.
\subsection{Proof of Theorem~\ref{thm:1:first:stage}}
\begin{proof}
For each iteration $t$, we have
\begin{align}
   \nonumber &\E[\Ls(\theta_{t+1}) - \Ls(\theta_t)] \\
   \nonumber\overset{(a)}{\le}& \E[\langle \nabla \Ls(\theta_t), \theta_{t+1} - \theta_t \rangle] + \frac{L}{2}\E[\|\theta_{t+1}-\theta_t\|^2]\\
   \nonumber\overset{(b)}{\le}& -\eta\left(1-\frac{\eta L}{2}\right) \E[\| \nabla \Ls(\theta_t)\|^2] + 2\eta^2G^2L\\
   \nonumber\overset{(c)}{\le} & -\frac{\eta}{2} \E[\| (1-\gamma)\nabla \Ls_a(\theta_t) + \gamma \nabla\Ls_b(\theta_t)\|^2] + 2\eta^2G^2L\\
   = & -\frac{\eta}{2} \E[(1-\gamma)^2\| \nabla \Ls_a(\theta_t)\|^2 + \gamma^2 \|  \nabla\Ls_b(\theta_t)\|^2 + 2(1-\gamma)\gamma\langle \nabla\Ls_a(\theta_t), \nabla\Ls_b(\theta_t)\rangle] + 2\eta^2G^2L, 
\end{align}
where (a) uses the smoothness of $\Ls$ in Assumption~\ref{ass:1} (ii); (b) uses $\theta_{t+1} = \theta_t - \eta g_t$ where $\E[g_t] = \nabla \Ls(\theta_t)$, and $\E[\|\Ls(\theta_t) - g_t\|^2]\le 4G^2$ due to Assumption~\ref{ass:1} (i) with Jensen's inequality; (c) uses $\eta L \le 1$ and the definition of $\Ls(\theta)$.
We consider tow cases. Case I. when $1>\gamma>\frac{1}{2}$, define $\alpha = \frac{1-\gamma}{2\gamma - 1}$, 
by Young's inequality we get
\begin{align*}
    &(1-\gamma)^2\|\nabla\Ls_a(\theta_t)\|^2 +   \gamma^2\|\nabla\Ls_b(\theta_t)\|^2 -2 \langle (1-\gamma)\nabla\Ls_a(\theta_t),  \gamma \nabla\Ls_b(\theta_t) \rangle \\
    \le &(1-\gamma)^2\|\nabla\Ls_a(\theta_t)\|^2 +   \gamma^2\|\nabla\Ls_b(\theta_t)\|^2 + \frac{1+\alpha}{\alpha} (1-\gamma)^2\|\nabla\Ls_a(\theta_t)\|^2 +  \frac{\alpha}{\alpha+1} \gamma^2\|\nabla\Ls_b(\theta_t)\|^2\\
    = &  (1-\gamma)\|\nabla\Ls_a(\theta_t)\|^2 +   \gamma\|\nabla\Ls_b(\theta_t)\|^2.    
\end{align*}
Case II. when $0<\gamma \le \frac{1}{2}$, define $\alpha = \frac{\gamma}{1-2\gamma}$, by Young's inequality we get
\begin{align*}
    &(1-\gamma)^2\|\nabla\Ls_a(\theta_t)\|^2 +   \gamma^2\|\nabla\Ls_b(\theta_t)\|^2 -2 \langle (1-\gamma)\nabla\Ls_a(\theta_t),  \gamma \nabla\Ls_b(\theta_t) \rangle \\
    \le &(1-\gamma)^2\|\nabla\Ls_a(\theta_t)\|^2 +   \gamma^2\|\nabla\Ls_b(\theta_t)\|^2 \frac{\alpha}{\alpha+1} (1-\gamma)^2\|\nabla\Ls_a(\theta_t)\|^2 +  \frac{\alpha+1}{\alpha} \gamma^2\|\nabla\Ls_b(\theta_t)\|^2\\
    = &  (1-\gamma)\|\nabla\Ls_a(\theta_t)\|^2 +   \gamma\|\nabla\Ls_b(\theta_t)\|^2.    
\end{align*}
Therefore, we get
\begin{align}
  \nonumber\E[\Ls(\theta_{t+1}) - \Ls(\theta_t)] \le&  -\frac{\eta}{2} \E[(1-\gamma) \| \nabla \Ls_a(\theta_t)\|^2 + \gamma \|  \nabla\Ls_b(\theta_t)\|^2 ] + 2\eta^2G^2L\\
  \le&  -\frac{\eta}{2} \E[2\mu_a(1-\gamma) (\Ls_a(\theta_t)-\Ls_a^{\text{min}}) + 2\mu_b \gamma  (\Ls_b(\theta_t)-\Ls_b^{\text{min}})] + 2\eta^2G^2L,
\end{align}
where the last inequality is due to PL-condition of $\Ls_a(\theta)$ and $\Ls_b(\theta)$. Let define $\tau = \min(2\mu_a, 2\mu_b)$, then 
\begin{align}
  \nonumber&\E[(1-\gamma)\Ls_a(\theta_{t+1}) + \gamma\Ls_b(\theta_{t+1}) -(1-\gamma) \Ls_a(\theta_t) - \gamma\Ls_b(\theta_t)] \\
  \le&  -\frac{\eta\tau}{2} \E[(1-\gamma) (\Ls_a(\theta_t)-\Ls_a^{\text{min}}) + \gamma   (\Ls_b(\theta_t)-\Ls_b^{\text{min}})] + 2\eta^2G^2L,
\end{align}
which implies
\begin{align}
  \nonumber&\E[(1-\gamma)(\Ls_a(\theta_{t+1})-\Ls_a^{\text{min}}) + \gamma\Ls_b(\theta_{t+1})] \\
  \le&  \left(1-\frac{\eta\tau}{2}\right) \E[(1-\gamma) (\Ls_a(\theta_t)-\Ls_a^{\text{min}}) + \gamma   (\Ls_b(\theta_t)-\Ls_b^{\text{min}})] + 2\eta^2G^2L.
\end{align}
We thus have
\begin{align}
  \E[(1-\gamma)(\Ls_a(\theta_{t})-\Ls_a^{\text{min}}) + \gamma\Ls_b(\theta_{t})] 
  \le  \left(1-\frac{\eta\tau}{2}\right)^t  \Ls(\theta_0) + \frac{4\eta G^2L}{\tau}.
\end{align}
Since $\eta = \frac{2}{\tau m}\log\left( \frac{m\tau^2\Ls(\theta_0)}{8G^2 L}\right)$ and $\Ls_a^{\text{min}}= 0$, we have
\begin{align}
\nonumber  \E[\Ls_a(\theta_{m})] \le&  \left(1-\frac{\eta\tau}{2}\right)^m  \frac{\Ls(\theta_0)}{1-\gamma} + \frac{4\eta G^2L}{\tau(1-\gamma)}\\
\nonumber  \le & \exp\left(-\frac{\eta\tau m}{2}\right)\frac{\Ls(\theta_0)}{1-\gamma} + \frac{4\eta G^2L}{\tau(1-\gamma)} \\
\nonumber= & \frac{8G^2 L}{m\tau^2(1-\gamma)}\left( 1+ \log\left( \frac{m\tau^2\Ls(\theta_0)}{8G^2 L}\right) \right)\\
\le & O\left( \frac{4G^2L\log(m)}{m\tau^2(1-\gamma)}\right).
\end{align}
\end{proof}

\subsection{Proof of Theorem~\ref{thm:main:1:max:norm}}
\begin{proof} 
Since we have
\[
\max_{\theta \in \Omega(\rh,s)} \|\nabla \Ls_a(\theta) - \nabla \Lh_a(\theta)\| = \max_{\theta \in \Omega(\rh,s)}\left\|\frac{1}{m}\sum_{i=1}^m (f(x_i^a;\theta) - y_i^a)\nabla f(x_i^a;\theta) - \nabla \Ls_a(\theta)\right\|,
\]
using the Bousquet inequality~\citep{koltchinskii2011oracle}, with appropriate definition of the function space, we have, with a probability $1 - \delta$
\begin{eqnarray*}
\lefteqn{\max\limits_{\theta \in \Omega(\rh,s)} \left\|\frac{1}{m G}\sum_{i=1}^m (f(x_i^a;\theta) - y_i^a)\nabla f(x_i^a;\theta) - \nabla \Ls_a(\theta)\right\|} \\
& \leq & 2\E\left[\max\limits_{\theta \in \Omega(\rh,s)} \left\|\frac{1}{m G}\sum_{i=1}^m (f(x_i^a;\theta) - y_i^a)\nabla_k f(x_i^a;\theta) - \nabla_k \Ls_a(\theta)\right\|\right] + \frac{4F\log(1/\delta)}{3m} + \sqrt{\frac{2\sigma_P^2}{m}\log\frac{1}{\delta}} \end{eqnarray*}
where
$\sigma_P^2 \leq \frac{1}{G^2}\max\limits_{\theta \in \Omega(\rh, s)}\E\left[(f(x_i^a;\theta) - y_i^a)^2|\nabla f(x_i^a;\theta)|^2\right] \leq \max_{\theta \in \Omega(\rh,s)}\E\left[(f(x^a;\theta) - y^a)^2\right]\le 2\Ls_a(\theta_1) := r^2$ with the Assumption~\ref{ass:1} that $\|\nabla f(x;\theta)\|\le G$.
Using the Berstein inequality, we have, with a probability $1 - \delta$, 
\[
\rh^2 \geq r^2 - \frac{2F^2}{3m}\log\frac{2}{\delta} - 2F\sqrt{\frac{\rh^2}{m}\log\frac{2}{\delta}}
\]
Using the assumption that $m \geq 64F^2\log(2/\delta)$, we have, with a probability $1 - \delta$
\[
r^2 \leq 2\rh^2 + \frac{4F^2}{3m}\log\frac{2}{\delta},  
\]
and therefore
\[
\sigma_P^2 \leq 2\rh^2 + \frac{4F^2}{3m}\log\frac{2}{\delta}.
\]
As a result, with a probability $1 - 2\delta$, we have
\begin{align*}
&\max\limits_{\theta \in \Omega(\rh,s)} \left|\frac{1}{m G}\sum_{i=1}^m (f(x_i^a;\theta) - y_i^a)\nabla f(x_i^a;\theta) - \nabla \Ls_a(\theta)\right| \\
 \leq & 2\E\left[\max\limits_{\theta \in \Omega(\rh,s)} \left|\frac{1}{m G}\sum_{i=1}^m (f(x_i^a;\theta) - y_i^a)\nabla_k f(x_i^a;\theta) - \nabla_k \Ls_a(\theta)\right|\right] + \frac{3F}{m}\log\frac{1}{\delta} + 2\rh \sqrt{\frac{2\log(1/\delta)}{m}} \\
\leq & 4\E_{x,\sigma}\left[\max\limits_{\theta \in \Omega(\rh,s)} \left|\frac{1}{m G}\sum_{i=1}^m \sigma_i (f(x_i^a;\theta) - y(x_i^a)) \nabla f(x_i^a;\theta)\right|\right] + \frac{4F\log(1/\delta)}{m} + 2\rh\sqrt{\frac{2\log(1/\delta)}{m}}.
\end{align*}
We then bound $\E_{x,\sigma}[\cdot]$. Using Klei-Rio bound~\citep{koltchinskii2011oracle}, with a probability $1 - 2\delta$, we have
\begin{eqnarray*}
\lefteqn{\E_{x,\sigma}\left[\max\limits_{\theta \in \Omega(\rh,s)} \left|\frac{1}{m}\sum_{i=1}^m \sigma_i (f(x_i^a;\theta) - y(x_i^a))\nabla f(x_i^a;\theta)\right|\right]} \\
& \leq & 2\E_{\sigma}\left[\max\limits_{\theta \in \Omega(\rh,s)} \left|\frac{1}{m}\sum_{i=1}^m \sigma_i (f(x_i^a;\theta) - y(x_i^a))\nabla f(x_i^a;\theta) \right|\right] + 2\rh\sqrt{\frac{\log(1/\delta)}{m}} + \frac{9F}{m}\log\frac{1}{\delta}
\end{eqnarray*}
We can bound the expectation term using Dudley entropy bound (Lemma A.5 of~\citep{bartlett2017spectrally}),i.e.
\begin{align}
\nonumber&\E_{\sigma}\left[\max\limits_{\theta \in \Omega(\rh,s)} \left|\frac{1}{m GF}\sum_{i=1}^m \sigma_i (f(x_i^a;\theta) - y(x_i^a))\nabla f(x_i^a;\theta) \right|\right] \\
\nonumber \leq & \E_{\sigma}\left[\max\limits_{\theta \in \Omega(\rh,s), h \in S_d} \left|\frac{1}{m GF}\sum_{i=1}^m \sigma_i (f(x_i^a;\theta) - y(x_i^a))\langle \nabla f(x_i^a;\theta), h\rangle \right|\right] \\
\leq & \inf\limits_{\alpha} \left(\frac{4\alpha}{\sqrt{m}} + \frac{12}{m}\int_\alpha^{\sqrt{m}}\sqrt{\log\N(\G(\rh,s);\epsilon)} d\epsilon \right)
\end{align}
where $\N(\G(r,s); \epsilon)$ is the covering number of proper $\epsilon$-net over $\G(r,s)$. Here $\G(r,s)$ is defined as
\[
\G(\rh,s) = \left\{v \in \R^m: |v| \leq \frac{\sqrt{m}\rh}{F}, |v|_1 \leq \frac{\sqrt{m s} \rh}{F} \right\}
\]
Using the covering number result for sparse vectors (Lemma 3.4 of~\citep{plan2013one}), we have
\[
    \log \N\left(\G(\rh,s);\frac{\sqrt{m}\rh}{F}\epsilon\right) \leq \frac{Cs}{\epsilon^2}\log\frac{2m}{s}
\]
and therefore
\[
    \log \N\left(\G(\rh,s);\epsilon\right) \leq \frac{Cm\rh^2s}{F^2\epsilon^2}\log\frac{2m}{s}
\]
Plugging into the Dudley entropy bound, we have
\[
\E_{\sigma}\left[\max\limits_{\theta \in \Omega(\rh,s)} \left|\frac{1}{m F}\sum_{i=1}^m \sigma_i (f(x_i^a;\theta) - y(x_i^a)) \right|\right] \leq \inf\limits_{\alpha}\left(\frac{4\alpha}{\sqrt{m}} + \frac{12\rh}{F}\sqrt{\frac{Cs}{m}\log\frac{2m}{s}}\log\frac{\sqrt{m}}{\alpha}\right)
\]
By choosing $\alpha$ as
\[
    \alpha = \frac{3\rh}{F}\sqrt{Cs\log\frac{2m}{s}},
\]
we have
\[
\E_{\sigma}\left[\max\limits_{\theta \in \Omega(\rh,s)} \left|\frac{1}{m}\sum_{i=1}^m \sigma_i (f(x_i^a;\theta) - y(x_i^a)) \right|\right] \leq 12\rh\sqrt{\frac{Cs}{m}\log\frac{2m}{s}}\left(1 + \frac{1}{2}\log\left(\frac{m}{9\sqrt{Cs\log(2m/s)}}\right)\right)
\]
Combining the above results together, with a probability $1 - 3\delta$, we prove the inequality (\ref{thm:main:1:max:norm:L:a}).
Similarly, we can bound $\max_{\theta \in \Omega(\rh, s)}|\nabla \Ls_b(\theta) - \nabla \Lh_b(\theta)|$.
\end{proof}

\subsection{Proof of Theorem~\ref{thm:2:after:first:stage}}
\begin{proof}
Using the standard analysis of optimization, we have
\begin{align}
   \nonumber &\E[\Ls(\theta'_{t+1}) - \Ls(\theta'_t)] \\
   \nonumber\overset{(a)}{\le}& 
   -\eta\E[\langle \nabla \Ls(\theta'_t), \widehat\g_t \rangle] + \frac{\eta^2L}{2}\E[\|\widehat\g_t\|^2]\\
   \nonumber\overset{(b)}{=}& \frac{\eta}{2}\E[\| \nabla \Ls(\theta'_t) - \nabla \Lh(\theta'_t) \|^2] - \frac{\eta(1-\eta L)}{2}\E[\| \nabla \Lh(\theta'_t) \|^2] - \frac{\eta}{2}\E[\| \nabla \Ls(\theta'_t) \|^2] + \frac{\eta^2L}{2}\E[\|\widehat\g_t - \nabla\Lh(\theta'_t)\|^2]\\
   \nonumber\overset{(c)}{\le} & \frac{\eta}{2}\E[\| \nabla \Ls(\theta'_t) - \nabla \Lh(\theta'_t) \|^2]  - \frac{\eta}{2}\E[\| \nabla \Ls(\theta'_t) \|^2] + \frac{\eta^2 \sigma^2L}{2}\\
  \overset{(d)}{\le} & \frac{\eta}{2}\E[(1-\gamma)\| \nabla \Ls_a(\theta'_t) - \nabla \Lh_a(\theta'_t) \|^2 + \gamma\| \nabla \Ls_b(\theta'_t) - \nabla \Lh_b(\theta'_t) \|^2]  - \frac{\eta}{2}\E[\| \nabla \Ls(\theta'_t) \|^2] + \frac{\eta^2 \sigma^2L}{2}, 
\end{align}
where (a) uses the smoothness of $\Ls$ and $\theta’_{t+1} = \theta‘_t - \eta g_t$; (b) uses $\E[g_t] = \nabla \Lh(\theta_t)$; (c) uses $\E[\|\Ls(\theta_t) - g_t\|^2]\le\sigma^2$ and $\eta \le \frac{1}{L}$; (d) uses the convexity of norm operation $\|\cdot\|^2$. Hence, we need to bound both $\| \nabla \Ls_a(\theta'_t) - \nabla \Lh_a(\theta'_t) \|$ and $\| \nabla \Ls_b(\theta'_t) - \nabla \Lh_b(\theta'_t) \|$. Since we assume
that $\theta'_t\in\Omega(\rh, s)$, it is thus sufficient to bound $\max_{\theta\in\Omega(\rh, s)}\| \nabla \Ls_a(\theta) - \nabla \Lh_a(\theta) \|$ and $\max_{\theta\in\Omega(\rh, s)}\| \nabla \Ls_b(\theta) - \nabla \Lh_b(\theta) \|$.

Plugging in the bounds for $\|\nabla \Ls_a(\theta_t') - \nabla \Lh_a(\theta_t')\|$ and $\|\nabla \Ls_b(\theta_t') - \nabla \Lh_b(\theta_t')\|$ in Theorem~\ref{thm:main:1:max:norm}, we have, with a probability $1 - \delta$,
\begin{align}
   \nonumber &\E[\Ls(\theta'_{t+1}) - \Ls(\theta'_t)] \\
   \le & \frac{C''\eta G^2}{n+m}\underbrace{\left(\log\left(\frac{6}{\delta}\right)+s\log\left(\frac{2\max(m,n)}{s}\right)\log(\max(m,n))\right)}\limits_{\Delta}\left(\rh^2 + F^2\right)  - \frac{\eta}{2}\E[\| \nabla \Ls(\theta'_t) \|^2] + \frac{\eta^2 \sigma^2L}{2}.
\end{align}
Using the similar analysis as the proof of Theorem~\ref{thm:1:first:stage}, after $t$ iterations, we have
\begin{align}
  \E[(1-\gamma)(\Ls_a(\theta'_{t})-\Ls_a^{\text{min}})] \le  \exp(-\eta\tau t)\Ls(\theta_1) + \frac{C'' G^2\Delta}{\tau(n+m)}\left(\rh^2 + F^2\right)   + \frac{4\eta G^2L}{2\tau}.
\end{align}
To ensure that the solution in the later iteration will fall into the range of $\Omega(\rh, s)$, we need the following conditions:
\begin{align*}
    &n+m \geq \frac{C''G^2\Delta (\rh^2 + F^2)}{8\tau\rh^2(1-\gamma)},\\
    & \eta =  \frac{\tau \rh^2(1-\gamma) }{16G^2L},\\
    & t \ge \frac{\log\left(\frac{8\Ls(\theta_1)}{\rh^2(1-\gamma)}\right)}{\eta\tau}.
\end{align*}
That is
\begin{align*}
    1-\gamma \geq\max\left\{ \frac{C''G^2\Delta (\rh^2 + F^2)}{8\tau\rh^2(n+m)}, \frac{16G^2L\log\left(\frac{\Ls(\theta_1)}{\rh^2(1-\gamma)}\right)}{\tau^2 \rh^2 t}\right\}.
\end{align*}
\end{proof}

\section{Conclusions}
Learning with noisily labeled data has been studied in may deep supervised learning tasks, and its two interesting phenomena called clean  data  first and  phase transition have been empirically observed. We provide a theoretical analysis for rethinking about these two empirical phenomena from the view of learning theory and non-convex optimization. The result reveals that the models first learn clean data and then after that the testing performance would not be degraded when the percentage of data with corrupted class labels is not too large.

\bibliographystyle{plainnat} 
\bibliography{ref}

\end{document}